# Survival of the flexible: explaining the recent dominance of nature-inspired optimization within a rapidly evolving world

James M. Whitacre

*Abstract*— Although researchers often comment on the rising popularity of nature-inspired meta-heuristics (NIM), there has been a paucity of data to directly support the claim that NIM are growing in prominence compared to other optimization techniques. This study presents evidence that the use of NIM is not only growing, but indeed appears to have surpassed mathematical optimization techniques (MOT) in several important metrics related to academic research activity (publication frequency) and commercial activity (patenting frequency). Motivated by these findings, this article discusses some of the possible origins of this growing popularity. I review different explanations for NIM popularity and discuss why some of these arguments remain unsatisfying. I argue that a compelling and comprehensive explanation should directly account for the manner in which most NIM success has actually been achieved, e.g. through hybridization and customization to different problem environments. By taking a problem lifecycle perspective, this paper offers a fresh look at the hypothesis that nature-inspired meta-heuristics derive much of their utility from being flexible. I discuss global trends within the business environments where optimization algorithms are applied and I speculate that highly flexible algorithm frameworks could become increasingly popular within our diverse and rapidly changing world.

## I. Introduction

An aim of this article is to explore the merits and limitations of two general approaches toward optimization that have been studied and developed over the last several decades. The first approach to optimization is labeled in this paper as Mathematical Optimization Techniques or MOT. This label is used in reference to a diverse set of highly successful programming techniques that are designed based on known (or assumed) mathematically formulated attributes of an optimization problem. Using well-grounded mathematical theory, MOT exploit these problem characteristics to search for solutions or construct solutions, sometimes with a guarantee of optimality. This approach includes well-known optimization sub-fields such as convex programming and linear programming as well as popular search space pruning techniques and many others listed in the Appendix.

MOT algorithms can be quite distinct from one another and they are applied across a diverse set of problems yet there are also unifying themes found in all MOT research. Most importantly, research in MOT closely follows the scientific tradition of decomposing a complicated topic into independent or separable parts. To be utilized, each class of MOT algorithms requires a set of corresponding conditions be met (e.g. linearity, convexity) or demand certain *a priori* knowledge about a problem.

A second approach to optimization is labeled in this paper as nature-inspired meta-heuristics or NIM. NIM algorithms have been developed using inspiration from natural systems that display problem-solving capabilities such as ants, swarm behavior, the immune system, genetic evolution, and non-inherited learning. The most popular NIM techniques have been applied to an extensive set of optimization problems including many that defy characterization within MOT-established problem classes. With the inspiration for each algorithm framework being unrelated to problem characteristics, it is common for these algorithms to be modified substantially when applied in practice. Implementation of these algorithms typically requires the integration of tools taken from MOT or other meta-heuristics and the incorporation of domain knowledge provided by subject matter experts [1]. So important is this hybridization/customization process that many NIM are referred to as Memetic Algorithms (cf. [2]); a term used in reference to the multi-scaled learning and environment-dependent ontogeny of multi-cellular life. Without a strong theoretical foundation or a clearly specified class of problems, it is less clear why particular NIM techniques are more popular than others, although it is implicitly assumed that the long-term persistence of a technique reflects some degree of sustained algorithm utility.

In the next section, I present findings from a study that evaluates optimization R&D intensity over the last several decades. Most notably, I present evidence within publication and patent trends that nature-inspired techniques are now more frequently used than algorithms falling within the classic domain of MOT. The remaining sections review and discuss possible reasons for this recent growth in popularity. Before evaluating research trends, it is worth commenting on the practical relevance of such a comparative study.

General comparisons between MOT and NIM are rarely considered in the literature, which may reflect a tendency to view these approaches as being applicable to different problems. Indeed, there are several instances (e.g. linear programming) where NIM and MOT are not strictly comparable based on the problems they have been designed to address. Yet from an applications perspective, it is also apparent that MOT and NIM are implemented on many of the same real-world problems and that they are frequently in competition for practical relevance. This competition is illustrated by the growing number of problems being solved today with NIM that would have previously been considered the strict domain of theoretically-grounded mathematical techniques, e.g. in scheduling, see Box 2, [3].

Our characterizations of MOT and NIM approaches are generalizations that are not perfectly reflected in all NIM or all MOT. However, to better understand the strengths and weaknesses of these general approaches to optimization, it is important for the purposes of this study to maintain distinctions between the approaches and thus to exclude algorithm classes that could blur these lines. With this in mind, this study does not consider meta-heuristics that are not inspired by nature. However, keyword validation tests presented in the Appendix indicate that grouping these other meta-heuristics with either NIM or MOT approaches would have little or no impact on the study's main findings (see Appendix).

## II. Trends on Algorithm Usage in Academia and Industry

Data on nature-inspired meta-heuristic usage in public, private, and academic sectors is sparse, however there has been some evidence that the use of NIM in computational problem solving is growing [4] [5] [6]. There is a steady stream of new nature-inspired algorithms being proposed, new journals and conferences being advertised, as well as a continuous supply of new applications being considered within academic research. In [5], bibliographic data on genetic algorithms is used to show that publications within this field experienced a 40% annual growth from 1978 to 1998. More recently a survey of evolutionary computation (EC) practitioners was conducted and found evidence that usage of EC techniques and jobs requiring EC expertise are growing at a super linear rate in both academia and industry [4]. Although these studies indicate increased use of evolutionary algorithms, it is not clear how these trends compare with similar research and development activity in other optimization techniques. To help address this knowledge gap, an analysis of publications and patent data from several publically accessible databases is provided for nature-inspired and mathematical optimization techniques in Box 1.

First, it is clear from the results in Box 1 that the usage of optimization techniques is increasing and that this is in many ways irrespective of the optimization paradigm considered. As discussed in greater detail in Section IV.B, there are likely a number of inter-related factors contributing to this growth including technological innovation, global prosperity, as well as growth in the number of problems that can be solved through computer-based methods, e.g. due to simulation technology and increased availability of computing resources. However, it is also apparent from Box 1 that metrics on the use of nature-inspired techniques have been increasing at a rate that is not matched by MOT. Moreover, the most recent data on publications, case studies, and patents strongly suggests that usage of nature-inspired techniques has now surpassed that of MOT. One aim of this paper is to attempt to understand why this is the case.

Before simply concluding that NIM's growing usage is a reflection of utility, it is important to consider alternative explanations. There are different reasons for growth within a competitive environment and not all of these are based on fitness. Here I comment on alternate explanations for growth based on historical bias, conceptual appeal, and other factors indirectly related to algorithm utility such as simplicity of use.

**Historical Bias:** In evolving complex systems, the prominence or activity of a component within a system can often be attributed to historical reasons. In particular, historical forces often bias growth in favor of past historical prominence, e.g. the well known "rich get richer" paradigm in economics and network science [7]. Comparing the rise in usage of the two optimization approaches in Box 1, it is apparent that historical arguments cannot account for the observed trends. Mathematical optimization techniques have a well-known rich history and were actively studied for decades prior to the first appearance of nature-inspired meta-heuristics. US patents of mathematical optimization techniques for solving linear programming and dynamic programming problems were first granted in 1972, while the first NIM (simulated annealing) was not patented until 1986. Taking the data from Figure 3a, for the ten years leading up to 1990 there were 2525 MOT journal publications compared with only 204 for NIM. Over the next ten years the relative size of this gap narrows (MOT=15509, NIM=8381), however the historical advantage at the turn of the century was still in MOT's favor.

**Conceptual Appeal:** NIM-biased growth might also occur for superficial reasons such as the conceptual appeal of a "nature-inspired" approach. Despite striving for objectivity in academic discourse, the infectivity of ideas in science is not always aligned with these ideals. In addition to support from observable, empirical and measurable evidence, the conceptual appeal or intuitive nature of an idea can play a role in determining its influence on scientific culture and scientific trends [8] [9]. Thus it is at least plausible that interest in NIM is partly due to an innate affinity for biological phenomena (e.g. because of their exceptional complexity, adaptability, and resilience) and a relatively lower affinity for rigorous mathematical theory.

While the conceptual appeal of nature-inspired approaches may influence academic publication trends, there are reasons to believe that conceptual appeal would have less of an influence on patent trends and industry usage. In a competitive marketplace, tradeoffs between cost and anticipated efficacy of an algorithm should play a considerable role in determining the utilization rates of services. That genetic algorithms are continually applied in industrial scheduling problems (see Box 2) can be seen as evidence of some degree of algorithm utility. Similar arguments can be applied to the patent trends (see Figure 1), which are a commonly used proxy for measuring commercial interests. In short, the costly decision to file for a patent is likely based on anticipated efficacy (or proven efficacy in the case of protective/defensive patenting) and not based on superficial appeal.

**Simplicity of Use**: Another plausible explanation for NIM popularity is their ease of use. The simplicity of implementation has been repeatedly mentioned in the literature as an advantage of nature-inspired optimization and this view is probably shared by most of its practitioners. Performance issues aside, it is often the case that NIM's can be quickly implemented with little effort, thus inertia should be less of an issue when industry end users consider implementing an NIM for the first time. However it is unclear whether such views are actually shared by industry stakeholders and end users. For instance, in Section 7 of [4] they survey Evolutionary Computation practitioners in industry about the challenges facing EC uptake and they report that some of the greatest perceived obstacles include that the algorithms are "too hard to apply" (14% of respondents) and are "poorly understood" (40% of respondents). In short, it is not clear whether the ease of implementation, perceived or actual, is a factor that influences NIM popularity outside academia.

To summarize, historical bias cannot account for the trends reported in this study, while conceptual appeal cannot be ruled out as a possible contributing factor. Assuming the presence of a competitive marketplace, it also appears plausible that industrial usage trends are at least partly indicative of algorithm utility. On the other hand, direct data of industry usage, success, and failure rates of NIM and MOT techniques is lacking and this is clearly needed to test the validity of these conjectures. Nonetheless, the trends shown in Box 1 are intriguing and raise questions about why nature-inspired algorithms are frequently used in contemporary optimization problems. The next section reviews past explanations of NIM utility. I then explore and expand upon the hypothesis that the flexibility of an algorithm framework is one of the most important factors determining the frequency of algorithm usage. Section 3 also reviews trends taking place in industry and society, which is used to make predictions about how algorithm utility will be evaluated in future optimization problems. A summary of the main findings and arguments is given in Section 5 with experimental methods provided in Section 6.

### III.   EXPLANATIONS OF ALGORITHM UTILITY

Early arguments in favor of NIM focused on fitness landscape features or theories related to the operation of genetic algorithms, e.g. schema theory [10] and the building block hypothesis [11]. For instance, genetic algorithms were often touted for their (relatively) low sensitivity to problem features such as discontinuities in the fitness landscape, non-Gaussian noise in objective function measurements, non-stationarity of a problem, errors in determining objective function gradients, and numerical rounding errors in computer calculations [12] [13] [14]. Their success in multi-objective and multimodal problems have also been commonly cited as advantages. Furthermore, many NIM involve population-based searches that readily benefit from distributed computing resources; something increasingly available in industry and academia. More recent studies have investigated the conditions where common NIM design features are beneficial such as population-based search [15], recombination [16], or the conditions when problems are difficult for a particular technique [17]. Others have aimed to derive a better understanding of NIM by investigating the search characteristics and convergence properties of an algorithm [18] [19] [20] [21] [22] [23] [24] [25].  Although many of these studies have provided theoretical insights, the studies are often by necessity restricted to specific algorithms and specific problems. This limits their explanatory power when attempting to understand the general propensity of NIM techniques to be applied across diverse applications.

When considering the practical challenges faced when applying an algorithm to a new problem, another narrative surrounding the success of NIM often arises based on the concept of algorithm flexibility. As is common knowledge by practitioners within the field, an NIM's success or failure depends upon the designer's ability to integrate domain knowledge into the algorithm design and generally customize the algorithm to handle the particular needs of the customer and problem. The importance of customization to the success of a GA has been thoroughly documented over the last 15 years within monographs, reviews and individual studies [26] [27] [28] [29] [30] [31] [32] [33] [34] [35] [36]. In the particular case of GA applied to industrial scheduling problems, which is reviewed in Box 2, it is not surprising that most successful case studies involve a custom GA or a memetic algorithm.

The notion of algorithm flexibility and its relation to utility is conceptually simple and is illustrated in Figure 6. In short, the utility of an algorithm, i.e. the ability to generate solutions that are useful to a client, is derived from the ability to adapt the algorithm to the unique attributes of a problem and not based on "off the shelf" performance characteristics. Adaptation in this context refers to the selective retention of sequential design changes that improve an algorithm's performance (e.g. final solution fitness) on a problem. Flexibility then is a reflection of relaxed design constraints that allow for the integration of customized routines that exploit problem knowledge.

## IV.   SURVIVAL OF THE FLEXIBLE AND THE ORIGINS OF ALGORITHM UTILITY

The idea that algorithm design flexibility is important to NIM utility is not new and in many ways is a simple and intuitive concept (e.g. [33] [32] [36]).  However, little effort has been devoted to exploring its theoretical basis or its practical implications for the field. For instance, few studies have explicitly considered when design flexibility is most relevant to algorithm utility in commercial settings or the consequences that an emphasis on flexibility would have for future algorithm research. The remainder of this paper explores some of these issues from the perspective of a problem's lifecycle.

Issues related to flexibility also arise in the study of dynamic optimization, e.g. see [37] [38]. While there are clear theoretical similarities between these topics, the current discussion focuses on changes in a problem that occur during its lifecycle or entirely new problem definitions in which sufficient algorithm modifications cannot (presently) be automated and instead require human creativity and insight, i.e. what is typically labeled as algorithm design changes. In short, this discussion considers changes in a problem definition that are broader than what would normally be considered as a non-stationary problem.

### A.  Relevant timescales in algorithm design adaptation

For many problems of practical interest, a search process will exhibit a trade-off between solution quality and the computational costs expended. Similarly, the performance of an adaptable algorithm framework is expected to display a trade-off between solution quality and the amount of time expended on algorithm design adaptation. To understand flexibility, it is thus necessary to account for the efficiency and efficacy of the design adaptation process (Figure 6b). As elaborated on in this section, efficiency becomes relevant when there are tight deadlines constraining algorithm development time or when the problem is susceptible to changes in definition (e.g. size, scope) within its lifecycle that necessitate changes in algorithm design.

To illustrate when design flexibility is relevant in the context of a problem lifecycle, I introduce three basic timescales: algorithm runtime (T1), algorithm development time (T2), and problem lifespan (T3). T1 measures the time needed to reach a stopping criteria during the search process, T2 measures the total time permitted to design an algorithm for a problem, and T3 measures the amount of time that a problem is relevant, e.g. to a client.

If T1 is small compared with the time required to make an algorithm design change, the primary concern in algorithm development should be to quickly discover a sequence of design changes that provide sufficient solution quality. Surprisingly, the performance of the initial algorithm design is not of tremendous importance in this context, so long as it can be adequately modified in the given time (T2). The magnitude of T2 influences perceptions toward sufficiency and speed of algorithm design adaptation. If short development times are preferred by a client or necessitated by a short problem lifespan (T3), then preference is given towards an algorithmic framework that can quickly adapt to new conditions, e.g. movement to the left in the bottom graph in Figure 5b.

Interpretations of T3 depend on whether a problem is solved once (e.g. most design problems) or is solved many times, e.g. in scheduling. If a problem is solved once and a solution can be reached at any time during the problem's lifespan, then T3 poses a straightforward constraint on the feasibility of a particular algorithm, e.g. T2 must be less than T3. When problems are repeatedly solved, algorithm utility might be naively measured by its fitness improvement over other algorithms multiplied by the time it is implemented, e.g. Δ{solution quality} x {T3-T2}. However when T3 is small, the importance placed on the early stages of implementation can be unexpectedly high, e.g. due to the importance of being "first to market" or avoiding bottlenecks within a larger project. In this case, the rapid design of sufficient algorithms can trump what would otherwise appear to be a more superior alternative [1]. In summary, T2 has a strong impact on an algorithm's utility for a client, especially when T3 is small.

### 1) Algorithm adaptation during and after development

The problem lifecycle has been described as having a lifespan over which problems are relevant and a time window when algorithm development takes place. What is observed in practice is often more complicated. Once we look closely at the individual components of a problem lifecycle, the importance of algorithm flexibility can become pervasive.

First, aspects of a problem (e.g. constraints, problem formulation, and even the underlying aims) can change over the course of an algorithm development project. The reasons for these changes are varied [1]. Particularly for new problems, it is common to learn more about the underlying nature of the problem, and consequently want to change the problem definition, as one develops ways to solve it. A client's true interests are rarely captured entirely by a well defined problem formulation and more likely involve a network of connected sub-problems and soft objectives that exist as tacit domain knowledge. Although frustrating to optimization service providers, early success during algorithm development can also breed a desire for change, e.g. a desire to expand the scope of the problem. However, it is worth noting that a change in the problem definition does not necessarily reflect poor planning or poor understanding by the client. Instead, well-regarded insights from management science suggest that these problem changes are likely a consequence of intelligent yet *boundedly rational* individuals attempting to make sense of a dynamic and complex world (cf [39] [40]). These insights suggest that changes to a problem during algorithm development are not always preventable and are likely to be a persistent feature in future optimization contexts.

Changes to a problem can also occur for reasons outside the control of the client and may take place after an algorithm is already implemented, e.g. see [13]. Typical reasons for this include unexpected changes in a market or in the internal operating conditions of a firm [1].

In summary, problems can change during and after the time allocated to algorithm development. When this occurs, an algorithm must effectively adapt and do so quickly enough to keep up with changing requirements, e.g. of a client during algorithm development or a market during algorithm implementation. Under these circumstances, an algorithmic approach whose suitability depends strongly upon the conditions set out in the original problem definition may find itself less able to adequately accommodate new unexpected conditions that arise.

*B.  An evolving aim in optimization research*

The accelerating pace of technological, organizational, and social change has been documented using several surrogate metrics for technological progress, innovation, and manufacturing activity [41] [42] [43] [44] [45] [46] [47] [48] [49]. The rate of change is often not steady however and analogies have been made between organizational, technological, and market changes and the punctuated equilibria of biological evolution, i.e. where periods of apparent stagnation are followed by abrupt change [50] [51].

Technological progress and diversification is particularly rapid in information technology [44] [45], energy sectors [43], biochemical and petrochemical industries [46], and some manufacturing sectors [47] [48] [49].  Although rapid expansion is not universally observed across industries, it is particularly acute in domains where optimization algorithms are often applied, i.e. those sensitive to technological innovations in IT.

Technological innovations can drive growth, however they can also be disruptive.  There is plenty of documented evidence that product lifecycles are shrinking and that this is resulting in a subsequent push for shorter product development times across many sectors [52] [53] [54] [55] [56] [57] [58] (although also see [59]). In response, research on the competitive advantage of firms has been placing considerable emphasis on the value derived from rapid product development, innovation speed and adaptation in changing markets.  This emphasis is largely organized around the study of "dynamic capabilities" and "time-based competition" [60] [61] [62] [63] [64] [65] [66] [67].  The changing business conditions just described are likely to have a direct bearing on the utility of future algorithm frameworks. A summary of these global trends and others that are relevant to optimization research is given in Figure 7.

First, the number of new optimization problems can be expected to increase within industries that are growing and diversifying in response to technological and social change. Viewed as an environment of expanding and diversifying resources, the algorithms most often implemented are likely to be those that can most quickly exploit these diverse resources. In other words, algorithmic paradigms that are the most flexible to new conditions and can contribute to an organization's "time-based" competitive advantage are more likely to be utilized.

The second major trend is one of growing volatility in extant problems; for many industries the anticipation of future conditions (e.g. in organizational capabilities, resources, markets, competitors) is becoming more uncertain. In addition, the problems that a firm wants solved today may only partially resemble the problems that arise in the future. In such volatile environments, the utility of an algorithm framework will not be derived from the ability to solve a static problem. Instead it will be the ability to adapt to changing problem conditions that is likely to define the success or failure in the optimization algorithms of tomorrow.

*C.  Comments on design flexibility in NIM and MOT*

The origins of algorithm flexibility are not well understood, however the concepts outlined in this paper and illustrated in Figure 6 are useful for entertaining possible explanations.

MOT research decomposes the world of optimization problems into mathematically tractable domains involving precise assumptions and well-defined problem classes. The rationale for this decomposition is straightforward; an algorithm can be applicable for any problems meeting the requisite conditions. It should be stressed that when such conditions are satisfied, MOT are often superior to NIM techniques. One drawback however is that many practical optimization problems are not strictly members of these problem classes.

Such issues may become especially relevant to algorithm utility for problems that are non-stationary during their lifecycle. Models of a problem can change if a client decides that the original problem definition is no longer satisfactory, e.g. due to changes in a marketplace, manufacturing facility, personnel, raw materials, weather conditions, etc. Because MOT suitability depends on a problem meeting specific conditions, an MOT not only places constraints on the current problem definition but also on how that problem definition can change over time. If changes to a problem definition tend to eliminate mathematical conditions exploited by the MOT, then this will limit MOT suitability.

This contrasts sharply with observations in NIM research. Because the applicability of nature-inspired meta-heuristics is not strictly determined by formal problem classes, a research culture has emerged where it is acceptable to evaluate algorithm success for specific problems instead of evaluating algorithm utility based on performance generality (although generality is still sometimes implied, see [68]). While this has created challenges in making unambiguous scientific advances within NIM research, it also has interesting implications for long-term developments in the field. For instance, because NIM algorithm design adaptation is commonplace, an algorithm framework's popularity must rely on multiple successes in different contexts involving different algorithm variants. Under these circumstances, long-term algorithm survival/popularity is less likely to reflect the performance of the canonical algorithm and instead more likely reflects success in algorithm design modification across problem contexts (see Figure 6b). In short, it is plausible that the most successful NIM are those that are most readily adapted to the unique attributes of a problem.

## V. Conclusions

Historically, optimization problems were not thought of as having an expiration date. However, waning are the days when an industrial optimization problem could be defined and studied for years without the problem changing. More and more in today's firms, new problems rapidly come into existence and existing problems change due to new and unexpected conditions. Solution quality will always be a primary concern, however the algorithm development time and an algorithm's capacity to accommodate new information and new conditions is expected to become an increasingly valued asset when addressing real-world optimization problems.

This paper reported evidence that nature-inspired meta-heuristics are becoming increasingly utilized for optimization problems in academia and industry. I propose that this growing dominance is partly due to an inherent flexibility that allows meta-heuristics to be efficiently and effectively modified to fit the characteristics of a problem. In a volatile and dynamic world, NIM popularity may have less to do with the efficacy of a particular set of algorithm designs on a particular set of problems and more to do with the ability of nature-inspired meta-heuristics (but also the people and culture surrounding their development) to incorporate domain knowledge quickly and to be advantageously combined with other methods.

## VI.  Appendix: Methods

### A.  Box 1 data analysis: selecting keywords

Keywords for nature-inspired meta-heuristics (NIM) and mathematical optimization techniques (MOT) were selected based on several considerations. A list of keywords was first compiled from active researchers within the respective disciplines. This list was then expanded through concept mapping services such as Google Sets (labs.google.com/sets) and Kartoo.com using prototypical examples for NIM (genetic algorithms, evolutionary computation, nature-inspired optimization), MOT (mathematical programming, nonlinear programming, linear programming), and optimization (operations research, optimization, decision theory). This resulted in roughly 20 keywords for NIM and 40 keywords for MOT. The lists were then culled to between 10 to 12 keywords per group based on the following considerations: i) some search engines could not handle search strings larger than 256 characters, ii) some keywords were common to both MOT and NIM research, iii) some keywords had significant meaning outside of optimization  and iv) some keywords were redundant for the purposes of this study since they were always co-listed with a more commonly used keyword. It should also be noted that some MOT keywords refer to classes of optimization problems, however these terms are used almost exclusively within the MOT research community and therefore provided effective classifiers for MOT data.

### 1)  Keyword validation

Validation of keywords was conducted with the following aims: i) to determine whether the keywords were adequate representations of each approach, ii) to determine the degree of overlap in the data retrieved for the two approaches and iii) to determine whether the removal of (non-nature-inspired) meta-heuristic search terms had a considerable influence on the results. Validation tests were conducted using Web of Science (Wos) Query 1 (described in Section VI.B).  The NIM and MOT keywords that were ultimately selected and used to obtain the results in Box 1 are labeled as N1 and M1, respectively.  Validation keyword sets for each approach are labeled as N2 and M2.  Keywords for meta-heuristics not inspired by nature are labeled as MH.

<u>Selected Keywords</u>

**NIM keywords (N1):** genetic algorithm, evolutionary computation, swarm optimization, ant colony optimization, memetic algorithm, genetic programming, simulated annealing, estimation of distribution algorithm, nature inspired algorithm, bio-inspired optimization, evolutionary strategies

**MOT keywords (M1):** mathematical programming, constraint programming, quadratic programming, quasi-Newton method, nonlinear programming, interior-point method, goal programming, integer programming, simplex method, branch and bound algorithm, linear programming, dynamic programming

<u>Alternate Keywords</u>

**Alternate NIM keywords (N2):** swarm intelligence, hyper-heuristics, adaptive operator selection, multi-meme algorithms, self generating algorithms, honey bees algorithm, differential evolution

**Alternate MOT keywords (M2):** branch-and-cut, exhaustive search, branch and price, convex programming, stochastic programming, quasi-concave programming

**Meta-heuristics not inspired by nature (MH):** greedy randomized adaptive search, great deluge, squeaky wheel optimization, tabu, harmony search, unit-walk, stochastic local search, iterated greedy algorithms, iterated local search, cross entropy method, extremal optimization, stochastic diffusion search, reactive search optimization, random-restart hill climbing, variable neighborhood search

<u>Keywords removed from all lists</u>

reinforcement learning, artificial neural networks, data mining, game theory, learning classifier systems, evolutionary programming, gene expression programming, artificial immune systems, polynomial optimization, parametric programming, geometric programming, non convex programming, gradient methods, numerical algorithms, deterministic global optimization, Lagrangian relaxation method, KKT condition, transportation method, cutting plane method, line search, Hungarian algorithm, penalty method, Barrier method, upper bounding techniques, combinatorial optimization, convex optimization, robust optimization, non-smooth optimization, fractional programming, separable programming, linearly constrained optimization, mixed integer linear programming, affine-scaling, duality, global convergence, complementarity problems

In Figure 1 and Table 1, the inclusion of additional keywords for MOT (M1+M2) and NIM (N1+N2) expand the total results retrieved from the database by 7% and 10%, respectively (calculation defined by F1 below). The data retrieved using NIM and MOT search strings was found to have a 3% overlap, indicating that the two sets of data classifiers are retrieving unique, non-overlapping data (calculation defined by F2 below). Keywords for meta-heuristics not inspired by nature (MH) are found to have little overlap with NIM (3%) or MOT (1%). Furthermore, adding MH results to NIM or MOT only expands these results marginally (8% and 7%, respectively) and thus is unlikely to have altered the main findings.

$$F_1(X_1, X_2) = \frac{\{(X_1 \cup X_2) - X_1\}}{X_1}$$

$$F_2(X_1, X_2) = \frac{\{X_1 \cap X_2\}}{\{X_1 \cup X_2\}}$$

**Table 1 Summary of relationships between keyword sets. Details related to data, keyword sets, and calculations is provided in the text and Figure 1.**

| N1 = 29225 | N2 = 3428 | M1 = 40469 | M2 = 7834 | N1∩ M1 = 2151 |
|---|---|---|---|---|
| N1∩ N2 = 483 | N1∪ N2 = 32170 | M1∩ M2 = 5162 | M1∪ M2 = 43141 | N1∪ M1 = 67543 |
| $F_1$(N1,N2) = 10% | | $F_1$(M1,M2) = 6.6% | | $F_2$(N1,M1) = 3.2% |
| MH = 3415 | N1∩ MH = 1017 | N1∪ MH = 31623 | M1∩ MH = 533 | M1∪ MH = 43351 |
| | $F_1$(N1,MH) = 8.2% | $F_2$(N1,MH) = 3.2% | $F_1$(M1,MH) = 7.1% | $F_2$(M1,MH) = 1.2% |

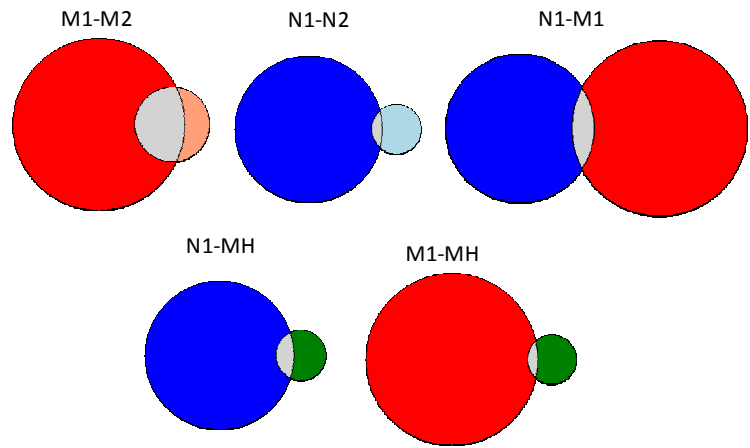

**Figure 1 Venn diagrams of data retrieved using WOS Query 1. Keywords for nature-inspired meta-heuristics (N1, N2), mathematical optimization techniques (M1, M2) and meta-heuristics not inspired by nature (MH) are defined in the text. Research trends reported in the paper are based on N1 and M1.**

*B. Box 1 data analysis: search engines*

*1) Delphion Patent Search*

Patent searches using Delphion (delphion.com) were restricted to granted US patents ("submitted only" patents were excluded). Each of the keywords were searched separately and only those contained in the "front pages" text of ten or more patents were included in the analysis (see lists below).

NIM: genetic algorithm, evolutionary computation, genetic programming, simulated annealing,

MOT: mathematical programming, constraint programming, quadratic programming, nonlinear programming, interior-point method, Integer programming, simplex method, linear programming,

*2) Google Scholar*

Google Scholar (scholar.google.com) was searched in one year increments to obtain time series data on publications containing one or more keywords. Specific categories (listed below) were excluded from the search if they were likely to include publication outlets directly associated with the field of optimization.

categories excluded

- Business, Administration, Finance, and Economics
- Engineering, Computer Science, and Mathematics

categories included:

- Biology, Life Sciences, and Environmental Science
- Chemistry and Materials Science
- Medicine, Pharmacology, and Veterinary Science
- Physics, Astronomy, and Planetary Science
- Social Sciences, Arts, and Humanities

*3) Web of Science (query one)*

Web of Science (WoS, http://apps.isiknowledge.com) was searched for all articles where the topic matched at least

one keyword. A publication frequency time series was extracted using the WoS results analysis tool. Citation databases that were searched are listed below (conferences databases were excluded).

Databases

- Science Citation Index Expanded (SCI-EXPANDED)--1945-present

- Social Sciences Citation Index (SSCI)--1956-present

- Arts & Humanities Citation Index (A&HCI)--1975-present

*4) Web of Science (query two)*

WoS query two used the same conditions as query one with the addition of the phrase "AND TS=(case study)".

*5) Scientific WebPlus*

Scientific WebPlus (http://scientific.thomsonwebplus.com) is a search engine that gathers a small selected set of websites that Thompson Scientific claims are most relevant to the search string. The website provides domain statistics associated with the returned results and these statistics were used in the analysis in Box 1.

*6) Google Trends*

Google Trends (www.google.com/trends) provides information on the relative search volume of individual keywords and phrases. More information on the analysis methods that are used is provided at (www.google.com/intl/en/trends/about.html). The search volume is measured by the Search Volume Index and not by absolute search volume. To calculate the Search Volume Index, Google Trends scales search volume data based on the first term entered so that the first term's average search volume over the selected time period is 1.0. Subsequent terms are scaled relative to the first term. Google Trends only allows 5 search terms to be compared at one time. To compare data for more than 5 terms required that all analysis be conducted starting with the same starting term (so that the data normalization that Google Scholar conducts is consistent). This was done with *genetic algorithms* as the first search term, however using this term only affects the scaling of the Search Volume Index and not the relative values reported. Search strings that returned negligible activity included: memetic algorithm, estimation of distribution algorithm, interior-point method, quasi-Newton method, goal programming, branch and bound algorithm.

## Box 1: Analysis of search algorithm usage

The following results evaluate research activity and usage of NIM and MOT algorithms. The analysis considers patent trends, publication trends, and trends related to internet sites and search traffic. Data for NIM and MOT is extracted using a set of 10 to 12 keywords related to each approach. See the methods section for more information on keyword selection, keyword validation, and how results were obtained.

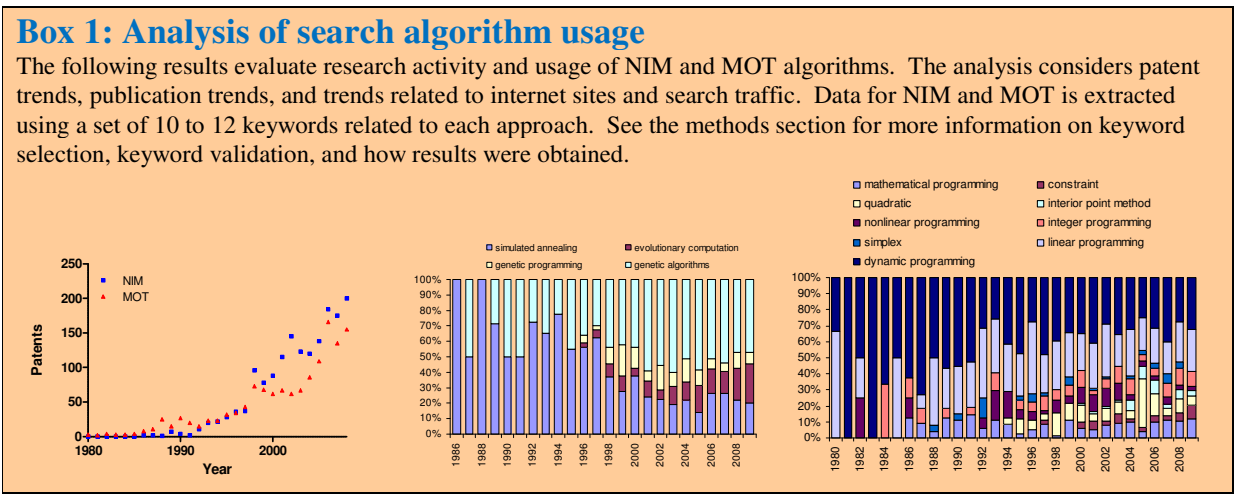

**Figure 2 US patent results: Patents granted in the US are shown for each year from 1980 to the present for all MOT and NIM related patents (left) with the relative contribution from individual keywords also presented for NIM (centre) and MOT (right). Only keywords occurring in the "front pages" of 10 or more US granted patents are included in the results shown in the centre and right graphs.**

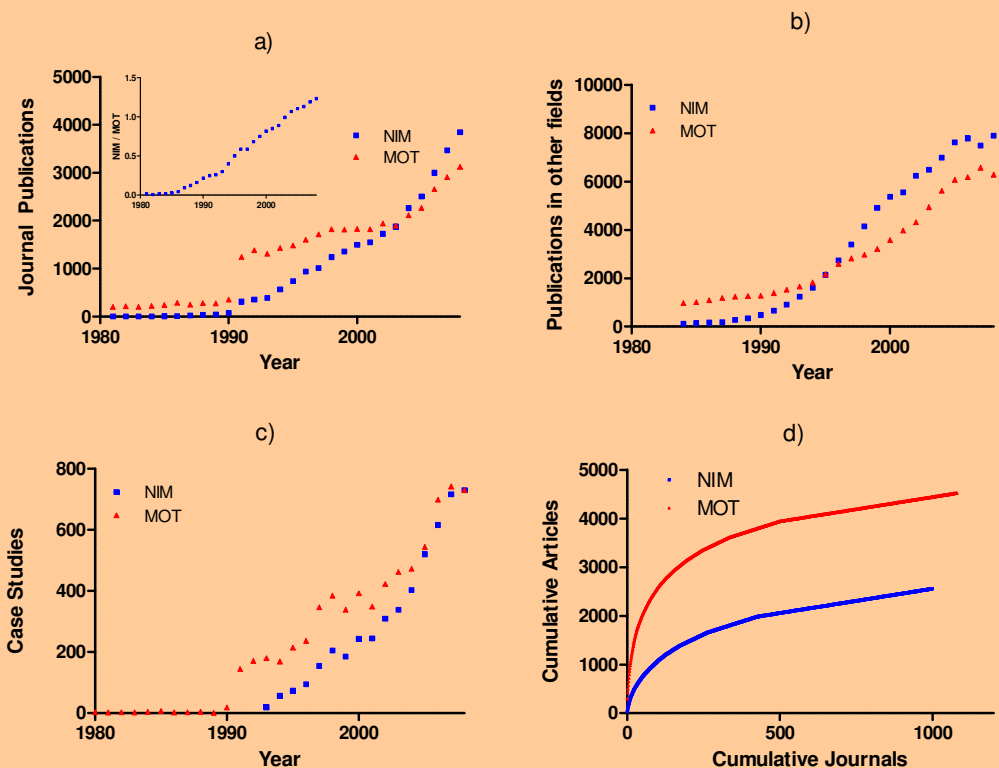

**Figure 3 Publication results: a) Journal publications found using Web of Science® (WoS). This publication time series appears to indicate three distinct trends, however the inset graph displays the annual NIM:MOT publication ratio, which exhibits a steady rise (from 1985-present). Since 2003, NIM has surpassed MOT in publication frequency and has grown an additional 158% compared with 71% for MOT. b) Publication data from Google Scholar is shown for any type of publication outlet (e.g. conferences, journals, books) but eliminates any publication sources that could be related directly to optimization research. For this data, the publication frequency of NIM surpassed MOT in the mid 1990s. c) Publication data from Web of Science is shown for any NIM or MOT publications that also include the topic "case study". For this data, NIM is only now approaching the same publication rate as MOT. d) Case study results from "graph c" are shown to indicate how publications are distributed across publication sources. These results demonstrate that per case study, NIM articles are spread out over a larger number of journals compared with MOT case studies. This relative diversity in publication outlets may indirectly indicate flexibility in applying NIM to distinct problems. For instance, a new NIM case study article had a 40% chance of being published in a new publication source compared with only 24% for MOT. Also, 60% of all MOT case studies can be found in just 11% of the journals where MOT case studies have been published. In contrast, 60% of all NIM case studies are found in 22% of the journals where NIM case studies have been published.**

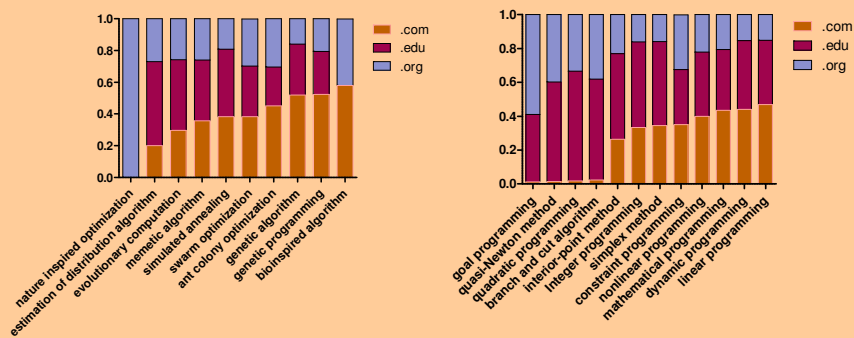

**Figure 4 Web domain statistics: Web domain statistics are shown for selected websites that have been determined by Scientific WebPlus® to have the most pertinent information related to the searched keyword. The goal with this analysis is to determine the extent that relevant content is located on academic versus non-academic websites.**

From this data, it was determined that the stated proportionalities for .edu domains are significantly different between NIM and MOT keywords, with MOT keywords more likely to have a higher proportion of .edu domain sites (t-test, unequal variances, p < 0.005).

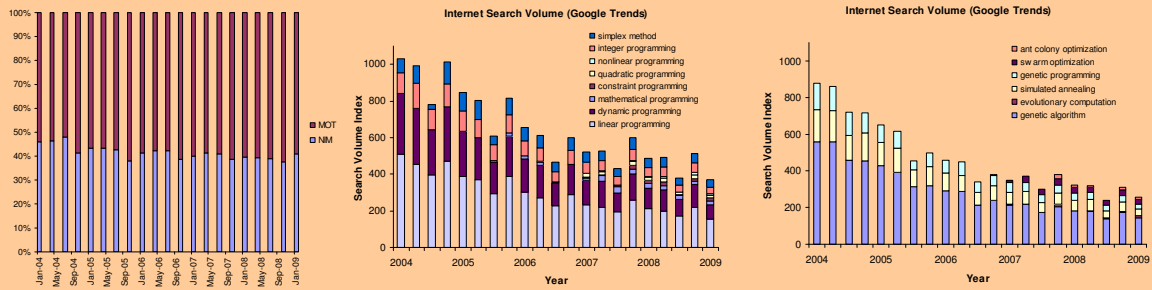

Figure 5: Internet search volume results: Google Trends® was used to evaluate internet search volume for each of the keywords from 2004 (the oldest recorded data) to the present. In presenting this data, the goal was to provide an indicator of overall interest in individual keywords related to MOT and NIM. The proportion of search volume related to MOT has remained approximately constant between 55% and 60% relative volume (left), however the search volume time series data for MOT (centre) and NIM (right) indicate a steady decline in internet search activity. In contrast with data from earlier figures, this could indicate that general interest in both MOT and NIM keywords is shrinking.

## Box 2: GA in industrial scheduling problems

This text box reviews recent evidence that GA's and GA hybrids are being successfully applied to industrial scheduling problems. I consider evidence based on: Table 2) case studies where government or industry is directly involved, Table 3) surveys of GA's applied to "industrial strength" test problems, and Table 4) scheduling optimization companies that utilize GAs or hybrids in their software.

**Table 2 GA and GA hybrids applied to scheduling problem case studies. The studies were selected from a WoS search for recent articles including the topics "case study" AND "industry" AND "genetic algorithm" AND "scheduling". Only studies with government or industry participation were included.**

| Company | Problem | Algorithm | Results | Year, Ref |
|---|---|---|---|---|
| General Electric | Maintenance scheduling for low earth orbit satellites | Custom GA | p.21, col.2: Found solution that was within 0.002% of the global optima in about 10 minutes of computation. | 2006, [33] |
| Chilean Foundry: Name Not Provided | Task scheduling in small Chilean foundry. | GA | p.4: Solutions better than an expert found (the production chief) for 50% of cases. | 2005, [67] |
| Dow AgroSciences | Scheduling of new products to market | Custom GA | p.7, tab.2: 18 projects in portfolio. None are solvable by DM. No MH beat all others on all projects, although GA was 10% better on average compared to second best MH. Authors estimate 10s of millions of dollars saved. | 2004, [68] |
| Far East Electro-Mechanical Manufacturing Plant: Name Not Provided | Scheduling design tasks in circuit board | Custom GA | p.8: Claim that good solutions are obtained, however benefits from GA are inconclusive since alternate methods are not compared. | 2004, [69] |
| Chinese company providing parts for internal combustion engines: Name Not Provided | Production scheduling | GA, Custom GA | p.18: 38% improvement in resource utilization. 31% improvement in completion time of jobs (compared to current operations). | 2005, [70] |
| PRECON S.A. (Spanish concrete company) | Scheduling production of prefabricated concrete parts. | Custom GA | p.14: 12% improvement in costs compared to current operations. | 2007, [71] |

**Table 3 Reviews and large scheduling problem studies.**

| Problem | Alg | Overview | Year, Ref |
|---|---|---|---|
| Resource-Constrained Project Scheduling | Custom GAs | This review provides evidence that GA hybrids are dominating scheduling optimization research. Evidence is based on artificial scheduling problems. | 2006, [67] |
| Airline Crew Scheduling | Custom GA | This paper considers airline crew scheduling with 28 real datasets taken from an airline. Problem sizes ranged from small to large however problem definitions appear to be simplified (e.g. less constraints) relative to other real world problems. GA reaches within 4% of global optimal solution on average. | 2004, [68] |
| Workforce Constrained Preventative Maintenance Scheduling | ES, SA | This paper looks first at Evolution Strategies (ES) on 60 small scale scheduling problems and shows ES can quickly reach optimal solutions. On 852 large scale problems (where optimal solutions are not known), ES was shown to find what they call "near optimal" solutions 12 times faster than Simulated Annealing (SA). | 2000, [69] |
| machine scheduling, timetabling, manpower scheduling, industrial planning. | MA | This paper surveys memetic algorithms applied to several types of artificial scheduling problems. Systematic comparisons with other types of algorithms are not available. However when taken as a whole, the authors claim that the studies surveyed provide convincing evidence that the MA algorithm framework has been highly useful for a range of scheduling problems. | 2007, [3] |

**Table 4 Private companies that provide schedule optimization services using a GA or GA-hybrid.  Most service providers also explicitly point out that algorithms are customized to fit particular client needs. The table is not exhaustive, e.g. companies such as OptTek Systems, Inc, AntOptima SA, are known to incorporate other NIMs into their algorithm designs.**

| Company | Online info | Description |
|---|---|---|
| Adaptive Intelligent Systems | http://www.adaptive-is.com | AIS provides custom solutions in scheduling and other domains using nature-inspired tools. They also provide consulting services to support clients with their own implementation of nature-inspired metaheuristics. |
| NuTech Solutions | http://nutechsolutions. com/index.asp http://nutechsolutions. com/lit_featured.asp | NuTech provides a range of nature-inspired optimization and computer modeling services. The second link describes their success with solving Air Liquide's scheduling distribution problems. |
| XpertRule | http://www.xpertrule.com/ pages/case_ud.htm | XpertRule is a GA optimization tool. The provided link describes a case study where XpertRule provided substantial improvements in distribution solutions for the company United Vinters and Distillers |
| Bio-Comp | www.bio-comp.com/industrial/ maximizeproduction.htm | Bio-Comp provides evolution-based solutions, some of which deal with scheduling related problems. |
| Advanced Computational Technologies | www.ad-comtech.co. uk/application.shtml www.ad-comtech.co. uk/study7.htm | ACT uses a GA to provide solutions to a number of problems including roster scheduling problems |
| Esteco | www.esteco.com/ schedulers.jsp | Esteco is a provider of optimization software.  Their optimization package includes two types of GA. |
| IcoSystem | http://icosystem.com /technology.htm http://icosystem.com /apps_personnel.htm | IcoSystem uses a GA for personnel management and scheduling problems |
| Bright Rivers | www.brightrivers.com /page.asp?id=home | Bright Rivers has developed a hybrid GA to solve scheduling optimization problems. |
| Blue Kaizen | www.bluekaizen.com/ | Blue Kaizen provides scheduling, planning, and routing solutions using a GA and other MH |
| SolveIt | www.solveitsoftware.com/ | SolveIt provides custom scheduling and planning solutions using a broad range of MH techniques including GA hybrids |

a) Classical understanding of
algorithm utility

b) Algorithm design flexibility as
a proxy for algorithm utility

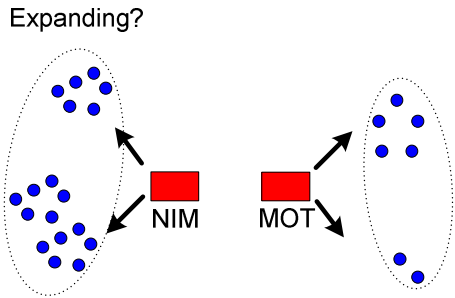

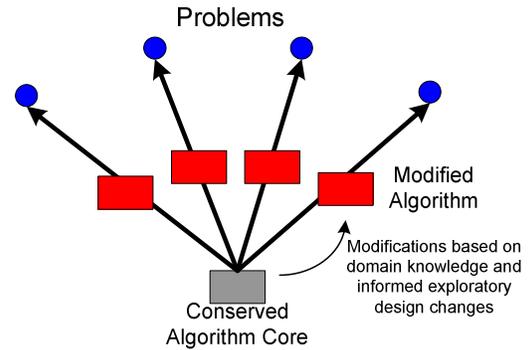

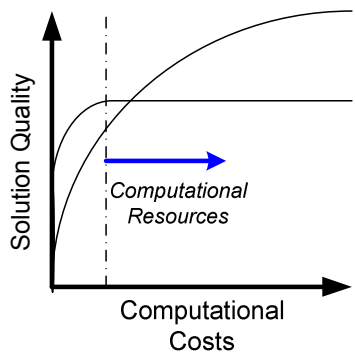

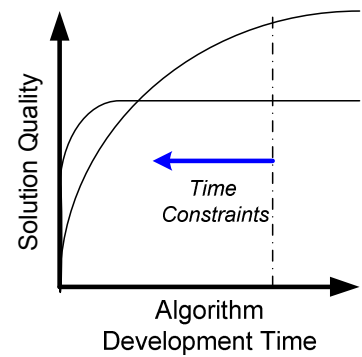

**Figure 6 Different perspectives for assessing algorithm utility.  a)** At small time scales, one only considers the utility of an algorithm from the perspective of a static algorithm design.  Under these conditions, classical explanations of algorithm utility can be  broken down into fitness landscape (top) and computational resource (bottom) arguments.  *Top*:  Popular algorithms are effective for problems with problem characteristics that are common, or are becoming increasingly so, for so called "real world" problems.  In the diagram, the distance between an algorithm (box) and problem (circle) indicates the suitability of the pairing. *Bottom*:  Algorithms often have different cost-benefit profiles, such that the preferred technique can sometimes depend on the amount of computational resources available.  The blue arrow indicates current trends in the availability of computational resources. **b)** From a broader perspective, this paper proposes that utility should be assessed based on an algorithm framework's adaptability towards different problems.  *Top*:  Illustration of how to assess the utility of an algorithm framework.  *Bottom*:  Algorithm adaptation profiles which show hypothetical values of solution quality as a function of the amount of time given to algorithm development. The blue arrow indicates current trends in the available algorithm development times (see Section 3.2).

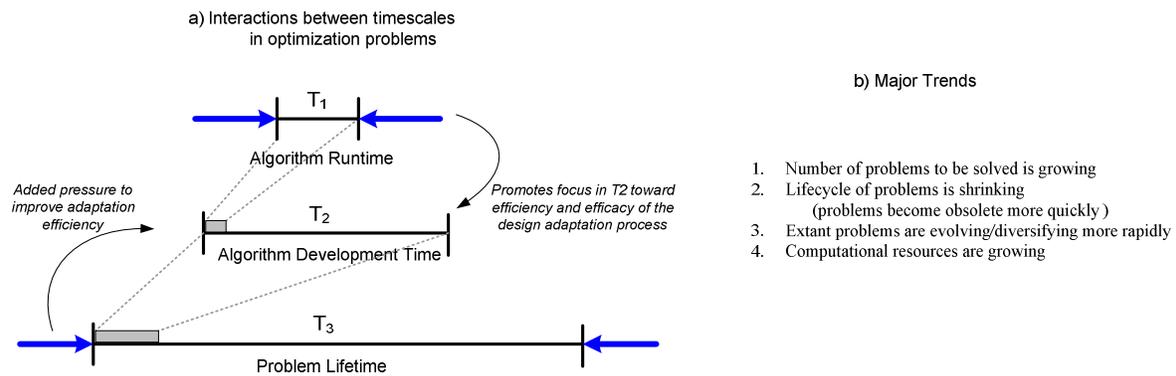

a) Interactions between timescales
in optimization problems

b) Major Trends

1. Number of problems to be solved is growing
2. Lifecycle of problems is shrinking
   (problems become obsolete more quickly )
3. Extant problems are evolving/diversifying more rapidly
4. Computational resources are growing

**Figure 7 a) Trends and interactions between optimization timescales. The amount of time needed for an algorithm to search for a solution ($T_1$) is decreasing due to technological improvements in computational resources. As a result, a smaller proportion of algorithm development time ($T_2$) is spent with the algorithm running and more is spent on algorithm design changes. Because a problem's lifespan ($T_3$) is decreasing and problem definition volatility is increasing, the available time to make algorithm design changes is becoming more constrained. b) Major trends that have a direct bearing on optimization research.**